\documentclass{sigchi}
\pdfoutput=1
\usepackage{balance}       % to better equalize the last page
\usepackage{graphics}      % for EPS, load graphicx instead 
\usepackage[T1]{fontenc}   % for umlauts and other diaeresis
\usepackage{txfonts}
\usepackage{mathptmx}
\usepackage[pdflang={en-US},pdftex]{hyperref}
\usepackage{color}
\usepackage{booktabs}
\usepackage{textcomp}
\usepackage{caption}
\usepackage{subcaption}
\usepackage{cuted}
\usepackage{capt-of}

\usepackage{microtype}        % Improved Tracking and Kerning
\usepackage{ccicons}          % Cite your images correctly!
\usepackage{todonotes}

\def\plaintitle{Multi-person Spatial Interaction in a Large Immersive Display Using Smartphones as Touchpads}

\def\emptyauthor{}
\def\plainkeywords{spatial intelligence; immersive spaces; interaction design; multi person interaction; smart phones}

\makeatletter
\def\url@leostyle{%
  \@ifundefined{selectfont}{
    \def\UrlFont{\sf}
  }{
    \def\UrlFont{\small\bf\ttfamily}
  }}
\makeatother
\def\pprw{8.5in}
\def\pprh{11in}

\definecolor{linkColor}{RGB}{6,125,233}
\hypersetup{%
  pdftitle={\plaintitle},
  pdfauthor={\emptyauthor},
  pdfkeywords={\plainkeywords},
  pdfdisplaydoctitle=true, % For Accessibility
  bookmarksnumbered,
  pdfstartview={FitH},
  colorlinks,
  citecolor=black,
  filecolor=black,
  linkcolor=black,
  urlcolor=linkColor,
  breaklinks=true,
  hypertexnames=false
}

\title{\plaintitle}

\numberofauthors{2}
\author{%
  \alignauthor{Gyanendra Sharma\\
    \affaddr{Troy, USA}\\
    \email{gyanendra.sharma870@gmail.com}}\\
  \alignauthor{Richard J. Radke\\
    \affaddr{Troy, USA}\\
    \email{rjradke@ecse.rpi.edu}}\\
}

\begin{document}
\maketitle
%\twocolumn[{%
%\renewcommand\twocolumn[1][]{#1}%
%\maketitle
%\begin{center}
%    \centering
%    \includegraphics[width=1.0\textwidth,height=3.0cm]{images/img5.jpg}[H]
%    \captionof{figure}{Multiple users simultaneously interacting with a large screen %using their smartphones and voices, coupled with spatial information about their physical %locations. This panoramic image shows a 5m tall 360-degree display wall with a 44m %perimeter. Please refer to the video hosted at \url{https://bit.ly/2GSapuM} for detailed %interaction examples.}
%    \vspace{10pt}
%    \label{fig:main}
%\end{center}%
%}]

\begin{strip}\centering
\includegraphics[width=1.0\textwidth,height=3.0cm]{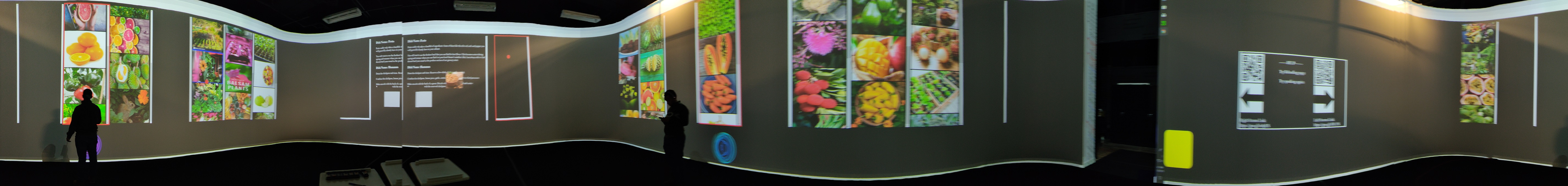}
\captionof{figure}{Multiple users simultaneously interacting with a large screen using their smartphones and voices, coupled with spatial information about their physical locations. This panoramic image shows a 5m tall 360-degree display wall with a 44m perimeter. Please refer to the video hosted at \url{https://bit.ly/2GSapuM} for detailed interaction examples.
\label{fig:main}}
%\vspace{10pt}
\end{strip}

\begin{abstract}
In this paper, we present a multi-user interaction interface for a large immersive space that supports simultaneous screen interactions by combining (1) user input via personal smartphones and Bluetooth microphones, (2) spatial tracking via an overhead array of Kinect sensors, and (3) WebSocket interfaces to a webpage running on the large screen.  Users are automatically, dynamically assigned personal and shared screen sub-spaces based on their tracked location with respect to the screen, and use a webpage on their personal smartphone for touchpad-type input.  We report user experiments using our interaction framework that involve image selection and placement tasks, with the ultimate goal of realizing display-wall environments as viable, interactive workspaces with natural multimodal interfaces.
    
\end{abstract}

\category{H.5.2.}{Information Interfaces and Presentation
  (e.g. HCI)}{User Interfaces} {}{}
  
%\category{H.5.2}{User Interfaces}{} \category{H.5}{Information Interfaces and Presentation (e.g., HCI)} %\category{H.5.1}{Information Interfaces and Presentation
%  (e.g. HCI)}{Multimedia Information Systems}{}{}


\section{Introduction}
Designing user interactive interfaces for large-scale immersive spaces  requires accommodations that go beyond conventional input mechanisms. In recent years, incorporating multi-layered modalities such as personal touchscreen devices, voice commands, and mid-air gestures have evolved as viable alternatives \cite{vogel2004interactive,malik2005interacting,kister2016multilens,bragdon2011code}. Especially in projector-based displays like the one discussed here, distant interaction via smartphone-like devices plays a pivotal role \cite{langner2019multiple}. 

Apart from the input modes of interaction, the size and scale of such spaces greatly benefit from contextualizing user locations within the space for interaction design purposes \cite{liu2014leveraging,wolf2016proxemic,kister2017grasp}. This is especially true for large enclosed displays such as CAVE \cite{cruz-neira,cruz-neira1}, CAVE2 \cite{febretti2013cave2}, CUBE \cite{rittenbruch2013cube} and CRAIVE \cite{gsharma}. Representing  physical user locations on such screen spaces presents considerable challenges due to spatial ambiguity compared to flat display walls.

In this paper, we present mechanisms for multiple users to simultaneously interact with a large immersive screen by incorporating three components: users' physical locations obtained from external range sensors, ubiquitous input devices such as smartphones and Bluetooth microphones, and automatic contextualization of personal vs.~shared screen areas. Discrete personal interaction regions appear on two sides of a rectangular enclosed screen, where users freely move to make spatial selections and manipulate or generate relevant images. The shared screen region between the two sides can be simultaneously used by multiple users to create a desired layout based on combinations of pre-selected text with user-curated images.

Our method and overall architecture allows multiple users to interface with the large visually immersive space in a natural way. Integrating personal devices and voice along with spatial intelligence to define personal and shared interaction areas opens avenues to use the space for applications such as classroom learning, collaboration, and game play. 

We designed controlled laboratory experiments with 14 participants to test the usability, intuitiveness and comfort of this multimodal, multi-user-to-large-screen interaction interface. Based on the results, we observe that the designed mechanism is easy to use and adds a degree of fun and enjoyment to users while in the space.

\section{Background and Related Work}

Our system is inspired by a diverse body of prior work, generally related to spatial sense-making in large immersive spaces, personal vs.~shared spaces in large screens, multi-user support, and interactions using ubiquitous devices such as smartphones.

\subsubsection{Spatial intelligence in immersive spaces}
Microsoft Kinects and similar 3-D sensors have been widely used for user locations or gestural interpretation in the context of various large screens \cite{ackad2015wild,yoo2015dwell,ackad2016skeletons} and common spaces \cite{ballendat2010proxemic}. Research has primarily been focused on developing mid-air gestures and other interaction mechanisms using methods similar to ray-casting, which require knowledge of spatial layout and users' physical locations \cite{kopper2008increasing}. A unique aspect of our system is the overhead Kinect  array that allows many users to be simultaneously tracked and their locations to be correlated to screen coordinates and workspaces.

\subsubsection{Personal vs.~shared spaces}

In terms of demarcating public vs.~personal spaces within large screens, Vogel and Balakrishnan \cite{vogel2004interactive} discussed how public displays can accommodate and transition between public and personal interaction modes based on several factors. This thread of research extends to privacy-supporting infrastructure and technologies \cite{brudy2014anyone,hawkey2005proximity}. Wallace et al.~\cite{wallace2017subtle} recently studied approaches to defining personal spaces in the context of a large touch screen display, which we cannot directly incorporate in our system but inspired our design considerations.

\subsubsection{Multi-user support}
Realizing large immersive spaces as purposeful collaboration spaces through multi-user interaction support remains an active area of research \cite{anslow2016collaboration}. Various approaches such as visualization of group interaction \cite{von2017giant}, agile team collaboration \cite{kropp2017enhancing}, along with use cases such as board meeting scenarios \cite{horak2016presenting}, have been proposed.  The Collaborative Newspaper by Lander et al.~\cite{lander2015collaborative} and Wordster by Luojus et al.~\cite{luojus2013wordster} showed how multiple users can interact at the same time with a large display. Doshi et al.~presented a multi-user application for conference scheduling using digital ``sticky notes'' on a large screen \cite{doshi2017stickyschedule}.

\subsubsection{Smartphones as interaction devices}

The limitations of conventional input devices for natural interactions with pervasive displays have led to several innovations, for example allowing ubiquitous devices such as smartphones to be used as interaction devices. Such touchscreen devices allow for greater flexibility and diversity in how interaction mechanisms with pervasive displays are materialized. Earlier concepts such as the one proposed by Ballagas et al.~\cite{ballagas2006smart} have evolved towards more native web-based or standalone application-based interfaces. For instance, Baldauf et al.~developed a web-based remote control to interact with public screens called \textit{ATREUS} \cite{baldauf2016your}. Beyond the touchscreen element of smartphones, researchers have investigated combining touch and air gestures \cite{chenAirTouch}, 3D interaction mechanisms \cite{du2011tilt} and using built-in flashlights for interaction \cite{shirazi2009flashlight}. 

\section{system Design}

Our system was designed and implemented in a large immersive display wall environment with a 5m tall 360-degree front-projected screen enclosing a 12m $\times$ 10m walkable area.  The screen is equipped with 8 $1200\times1920$ resolution projectors, resulting in an effective 1200 $\times$ 14500 pixel display, and contains a network of 6 overhead Kinect sensors for visual tracking of multiple participants.  %\rnote{I'm wondering whether naming CRAIVE raises an issue with blind review.}

\subsection{Spatial sense-making}
Large immersive spaces have exciting potential to support simultaneous multi-user interactions.  Flat 2D displays can support such functions simply by using multiple input devices with minimal consideration for physical user locations. However, to instrument large immersive environments for multi-person usage, it is necessary to demarcate personal vs.~collaborative or shared sub-spaces within the context of the large screen. Contextualizing physical user locations in the space plays an important role.

\begin{figure}[h!]
    \centering
    \includegraphics[width=0.8\linewidth]{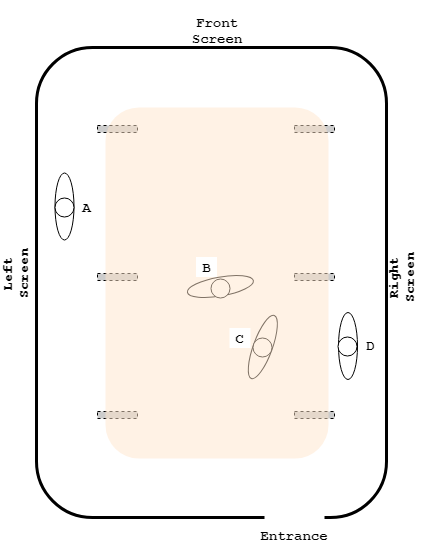}
    \caption{Users A and D are able to interact with the screen whereas users beyond 2 meters distance to the screen i.e users B and C are in the inactive region, represented in light red and thus cannot interact with the screen.}
\label{fig:active}
\end{figure}

To allow multiple users to interact with the screen at the same time, the large screen is subdivided into dynamic sub-spaces based on physical user locations.  The existing ceiling-mounted Kinect tracking system returns the \textit{(x,y)} location of each user in a coordinate system aligned to the rectangular floor space.  Although users are tracked wherever they are in the space, we enabled display interactions only for users that are located within 2 meters of the screen, as shown in Figure~\ref{fig:active}.  In this way, the center of the room acts as an inactive zone, where users can look around and decide on their next steps instead of actively participating at all times.   

In order to make this behavior clear to the users, we carefully calibrated the floor \textit{(x,y)} positions to corresponding screen locations. A key element of our design is a continuous visual feedback mechanism shown at the bottom of the screen when a user is in range, appearing as animated circular rings, as shown in Figures \ref{fig:s2}, \ref{fig:s3} and \ref{fig:s4}. This feedback serves a two-fold purpose. It makes users aware that their movements and physical locations are being automatically interpreted by the system, and it also allows them to adjust their movements to accomplish interactions with small sub-screens or columns on the large screen. Beyond the continuous feedback, we create discrete interaction spaces on the screen that change dynamically based on user locations. Thus, at a given point in time, users are able to visualize how the system continuously interprets their physical locations in real time, and also the column or sub-screen with which they are able to interact.

\subsection{Input Modes of Interaction}

We experimentally explored the viability of various input methods for multiple users to interact with the large screen.  These included the Leap Motion device for sensing mid-air gestures (which users found fatiguing and cumbersome to use), a fully voice-driven system (which had difficulty with some users' accents, and was discouraged by some recent studies \cite{nutsi2015multi,nutsi2015usability,sarabadani2018automatic}), and a smartwatch interface (which proved too small to easily control the large screen).  Ultimately, as described below, we use each user's own smartphone as a touchpad to control the large screen, which is both familiar and intuitive to use and has an immediate personal connection.

We developed a web application that can run on any touchscreen device connected to the internet.  Upon entering the environment, users showed their smartphone a QR code leading to the webpage. The webpage was designed to run as a trackpad, where familiar touch screen gestures such as tap, swipe, scroll, double tap, pinch, drag, and zoom were supported. Developing on a web platform removed the cumbersome process of users having to download and install a standalone application.  

\subsection{System Architecture}

\begin{figure}[h!]
    \centering
    \includegraphics[width=1.0\linewidth]{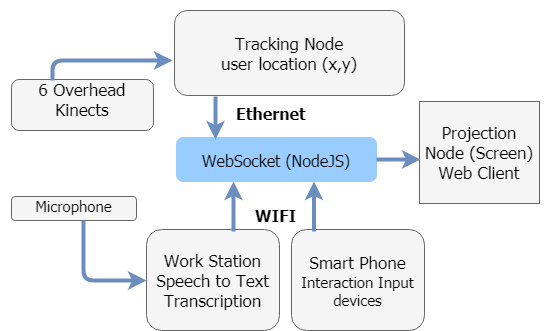}
    \caption{Overall system architecture.}
    \label{fig:sysarch}
\end{figure}

The system architecture of the overall system is shown in Figure \ref{fig:sysarch}.  It is primarily comprised of 3 components: (1) user input via smartphone and Bluetooth microphone, (2) spatial tracking via overhead Kinect sensors, and (3) the webpage running on the large immersive screen for visualization and output.  All components communicate with each other in real time using the WebSocket protocol. 
The users' smartphone gestures are sent via WebSocket to the web application running on the large screen, as well as any voice input, which is passed through the Google speech-to-text transcription service. The user tracking system is located in a different node, which sends the \textit{(x,y)} location of all users to the screen. The web application running on the large screen receives all the data, and displays dynamic feedback and visualizations accordingly.

\subsection{Overall System}
\begin{figure*}
\centering
\begin{subfigure}{.24\textwidth}
  \centering
  \includegraphics[width=0.95\linewidth]{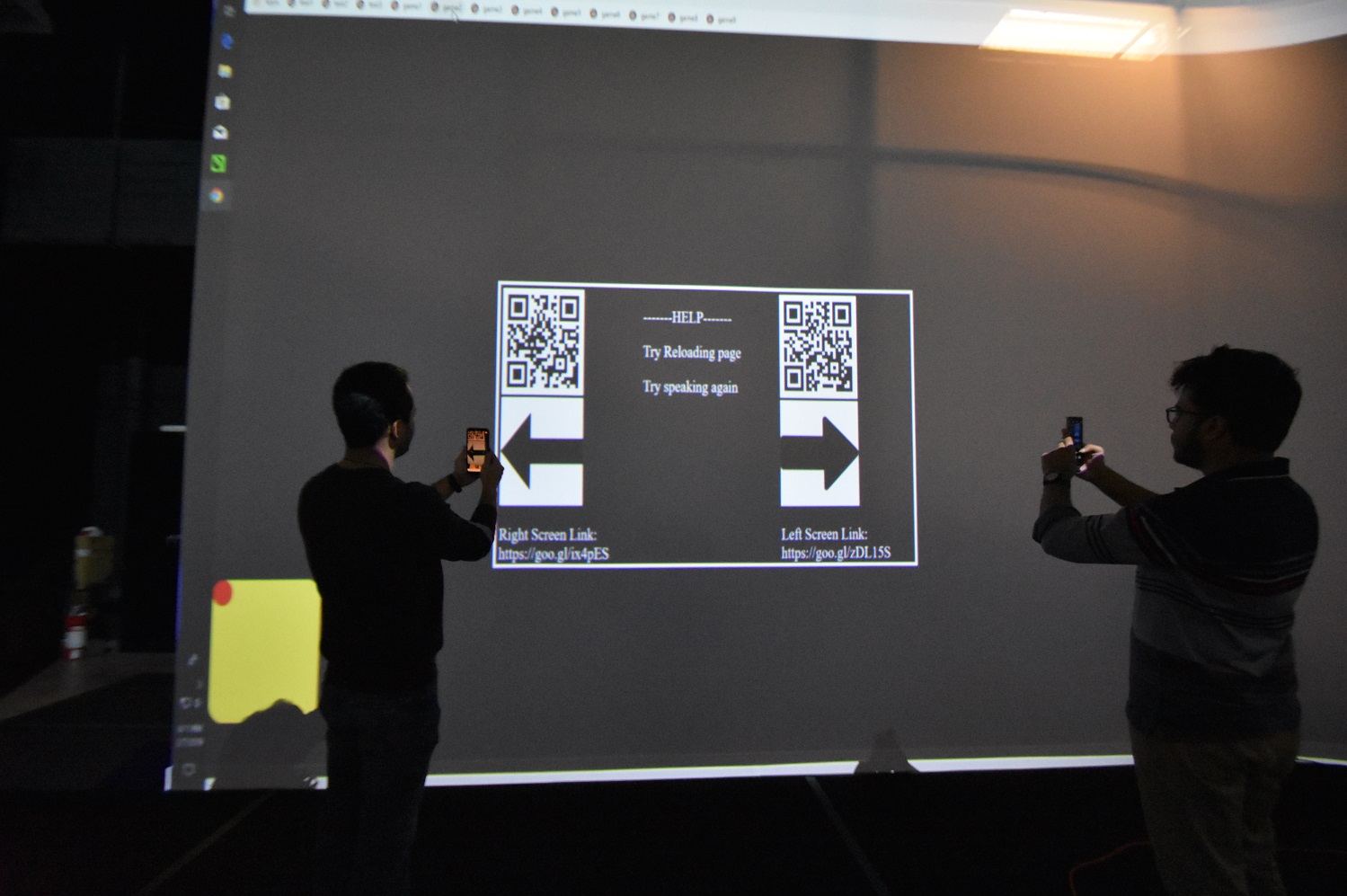}
  \caption{}
  \label{fig:s1}
\end{subfigure}%
\begin{subfigure}{.24\textwidth}
  \centering
  \includegraphics[width=0.95\linewidth]{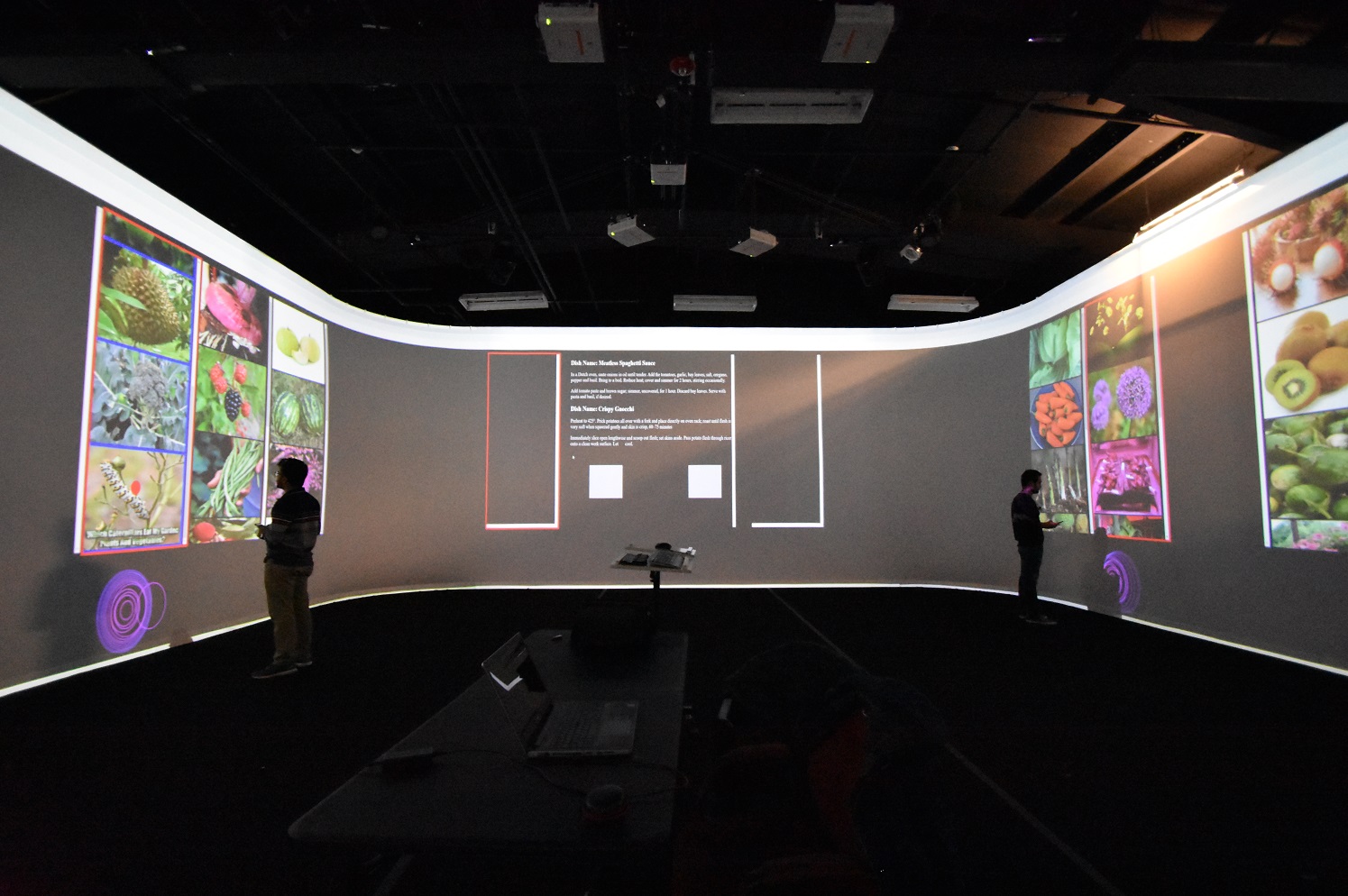}
  \caption{}
  \label{fig:s2}
\end{subfigure}
\begin{subfigure}{.24\textwidth}
  \centering
  \includegraphics[width=0.95\linewidth]{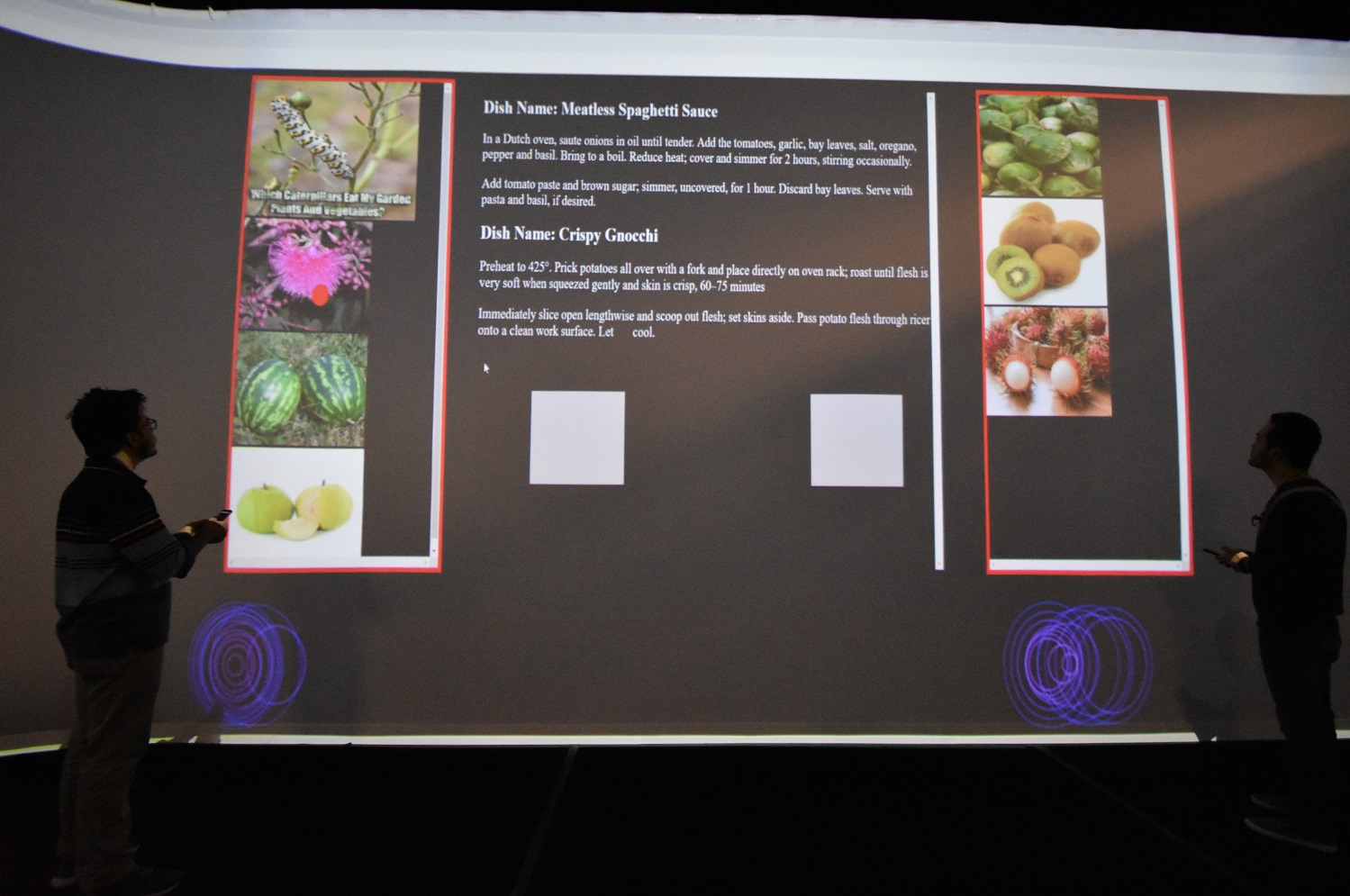}
  \caption{}
  \label{fig:s3}
\end{subfigure}%
\begin{subfigure}{.24\textwidth}
  \centering
  \includegraphics[width=0.95\linewidth]{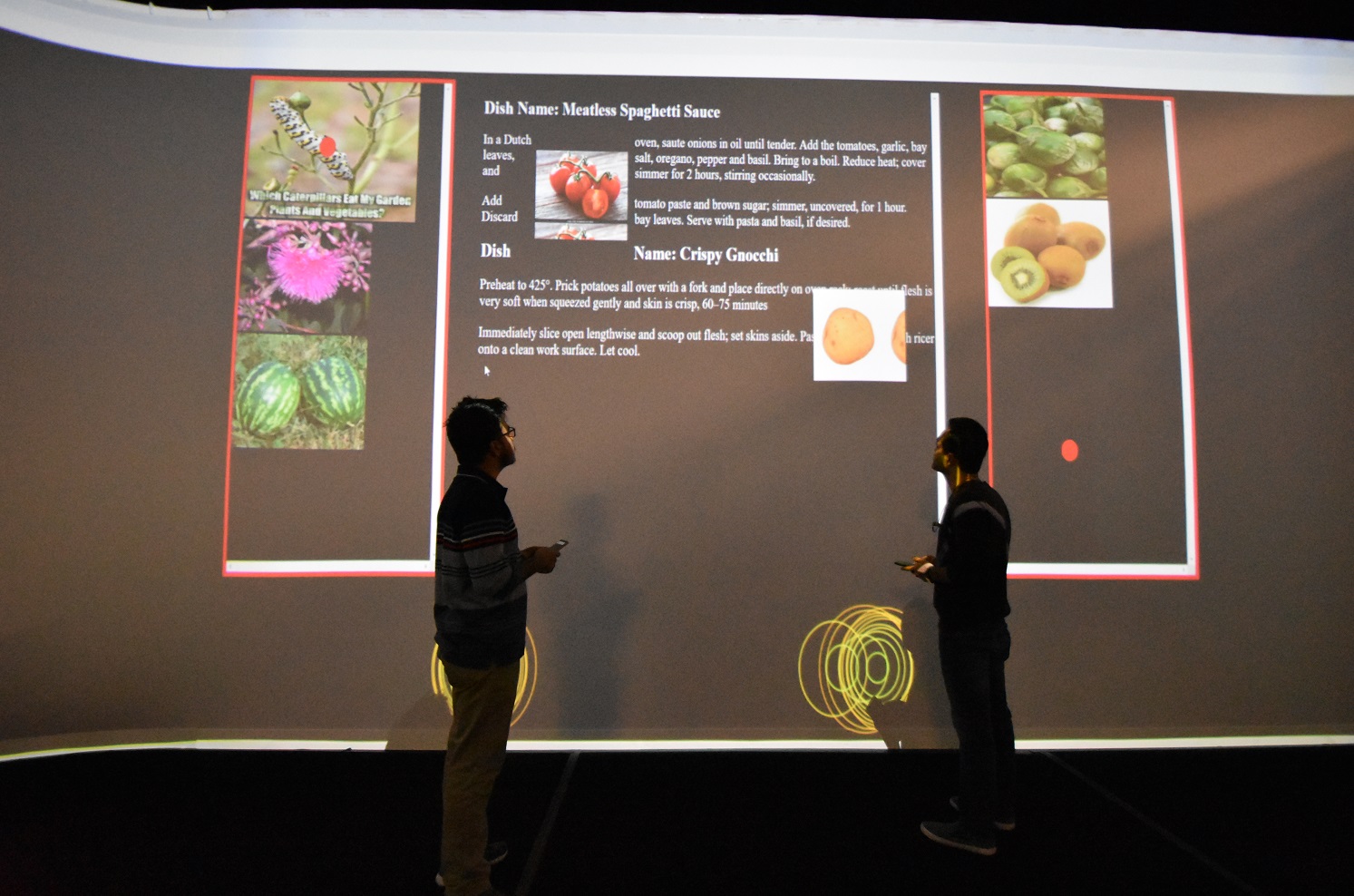}
  \caption{}
  \label{fig:s4}
\end{subfigure}
\caption{Multiple users during an experiment. (a) Each user scans a QR code. (b) Users work in their personal spaces using their smartphones and/or voice control. (c) Users move to the front screen to view their curated list of images. (d) Both users using the shared space to complete a full task.  For a better visualization, please refer to the video in the supplemental material.}% or anonymously hosted at \url{https://bit.ly/2GSapuM}.}
\label{fig:test}
\end{figure*}

Combining all the components discussed in the previous sections, we designed a multi-user spatial interaction mechanism for the large immersive space, using smartphones as input interaction devices and voice control for content generation. As shown in Figure \ref{fig:test}, two users can walk into the space and scan a QR code located near the entrance to launch the web application on their personal devices. The two QR codes correspond to the left and right sides of the big screen. As the users move towards their respective screens and come within the defined threshold of 2 meters, the location feedback and interaction mechanisms are activated, allowing them to interact with the individual columns as they see fit. We populate each of the columns with random images from the public Flickr API. As users move around the space, the continuous spatial feedback appears at the bottom of the large screen and the column with which each user can interact is highlighted in bright red.  Since the interaction column is tied to the spatial location of the user, they can be viewed as exclusive or personal to the user standing in front of it.

\begin{table}[h!]
\begin{tabular}{ll}
\hline
Phone Gestures & Screen Result \\ \hline \hline
Move & \begin{tabular}[c]{@{}l@{}}Move red pointer/drag image \\ on shared screen\end{tabular} \\
Tap & Select image \\
Swipe (Left or Right) & Move image/s to front screen \\
Swipe (Up or Down) & Scroll up/down personal column \\
Pinch & Shrink selected image \\
Zoom & Enlarge selected image \\
Double Tap & Enlarge/shrink selected image \\
Long Tap & \begin{tabular}[c]{@{}l@{}}Activate/deactivate drag \\ on shared screen\end{tabular}\\ \hline
Other input &  Screen Result\\ \hline \hline
\begin{tabular}[c]{@{}l@{}}Move (Physical user \\ locations)\end{tabular} & \begin{tabular}[c]{@{}l@{}}Select different column/\\ continuous circular \\ visualization\end{tabular} \\
\begin{tabular}[c]{@{}l@{}}Voice input ("Show \\ me pictures of X")\end{tabular} & \begin{tabular}[c]{@{}l@{}}Populate column with \\ pictures of "X"\end{tabular}
\end{tabular}
\caption{List of phone gestures, physical user movements and voice inputs, and their corresponding screen results.}
\label{tab1}
\end{table}

A red cursor dot that appears on the spatially selected column can be moved using the web application on the phone and acts similar to a mouse pointer. Table \ref{tab1} shows the list of supported gestures and how they translate to the big screen. Images that users select on the left and right screens can be moved to the front screen, which supports a personal column for each user as well as a large shared usage area. Users can move their personally curated images to the shared area on the front screen. In our particular case, we designed an application in which users can simultaneously drag their personal images around the shared screen to design a simple newspaper-article-like layout.

\section{User Studies}

We gathered 14 participants to test the usability of and gain feedback about our overall system. Only 3 participants had extensive prior experience with working in immersive display environments. We designed two experiments.  The first experiment was designed to gain a quantitative understanding of how long individual users take to perform various tasks on the screen using our system. The second experiment was designed as a simple game, where two users simultaneously work using both their personal and shared screens to come up with a final correct layout. This was largely designed to understand how comfortable users felt in the space and how intuitive the felt the system to be. For this experiment, users were mostly left on their own to complete the tasks based on their understanding of how the system works.

\subsection{Experiment 1}

Individual users were directed to use only the left screen, where they were asked to complete tasks based on the prompts appearing on the large screen. There were 9 columns, each filled with random images. Screen prompts would appear randomly on any of the 9 columns asking the user to complete various tasks, one after the other.  We tested all the gestures and inputs by asking the user to perform tasks shown in the second column of Table \ref{tab1}, except for the pinch, zoom, and double tap. Each task is completed once the user performs the correct input that corresponds to the displayed prompt.

For instance, if a user at a given point in time is in front of the 2\textsuperscript{nd} column, a prompt might appear in the 9\textsuperscript{th} column indicating \textit{``Select a picture from this column and move it to the front screen"}. Then, the user would physically move until the system highlights the 9\textsuperscript{th} column, and perform the corresponding scroll, tap, and swipe gestures. This would successfully complete the task and another prompt would appear on a different column, such as \textit{``Populate this column with pictures of dogs",} which would require a voice command. We recorded the time it took for the user to accomplish each task, including both the time it took to make spatial selections by moving between the columns and the time it took to successfully perform phone or voice input.

\subsection{Experiment 2}

We designed this experiment to be completed in pairs. Both users had completed Experiment 1 before taking part in this experiment. Our aim was to make sure users understand all input mechanisms and are comfortable to freely use the system.

On the front screen, where the shared screen is located, we presented a simple layout with two short paragraphs of text and image placeholders. Each paragraph consisted of a heading indicating a recipe name and text below describing the ingredients and preparation. Each user was responsible for finding an appropriate image for ``their'' recipe. Initially, the users independently move along the left and right sides of the screen, selecting one or more images and moving them to their personal columns on the front screen. Then, they move to the front screen and select the most likely candidate image from the refined set of images and move it to the shared screen with the recipe. A screen prompt on the large screen notifies the user whether a correct image was selected (i.e., a picture of the dominant ingredient in the recipe, such as an avocado picture for a guacamole recipe). Once the correct image is moved to the shared screen, users can perform a long-tap gesture on their phone to activate dragging on the shared screen. This allows the users to simultaneously drag their answer images to an appropriate location, which is generally next to the corresponding text. A screen prompt notifies the user once the target image has been moved to the required location on the shared screen. When both users complete their tasks on the shared screen, the full task is complete.

Each user pair was presented with 6 sets of recipe ``games''. 3 of the recipe pairs had the correct images already placed in one of the pre-populated columns and the users had to move around, scroll the columns, and locate the correct image. The other 3 pairs did not have the answer images in any of the columns and this required the users to generate content on their own by verbally requesting the system to populate a blank column with images of what they thought was the main ingredient in the recipe, one of which the user had to select and move to the front screen to verify.

We designed this setup to study whether users felt comfortable completing tasks based on the interaction mechanisms we designed for our display environment. We also wanted to find out if the users, most of whom had no prior experience with these kinds of spaces, found interacting with an unconventional immersive space such as this one to be fun and intuitive. Therefore, we asked the users to fill out a NASA-TLX questionnaire along with an additional questionnaire based on a 5 point Likert scale to obtain feedback on specific spatial, gestural, and voice input mechanisms that we designed.

\section{Results}

On average, each of the 14 participants performed 27 tasks during Experiment 1, where each of the 5 tasks appeared at random. Users were required to perform at least 20 and at most 35 tasks depending on the randomness of the distributed tasks as well as their speed at completing them. All tasks were assigned equal probability of appearing, except for voice control tasks, which appeared less often, according to the design considerations discussed earlier. The average number of tasks per user was distributed as follows: spatial selection (7.35), scrolling image columns (4.93), selecting an image (6.43), moving images to the center screen (6.65), and populating with voice (2.36).

\begin{figure}[htbp!]
    \centering
    \includegraphics[width=1.0\linewidth]{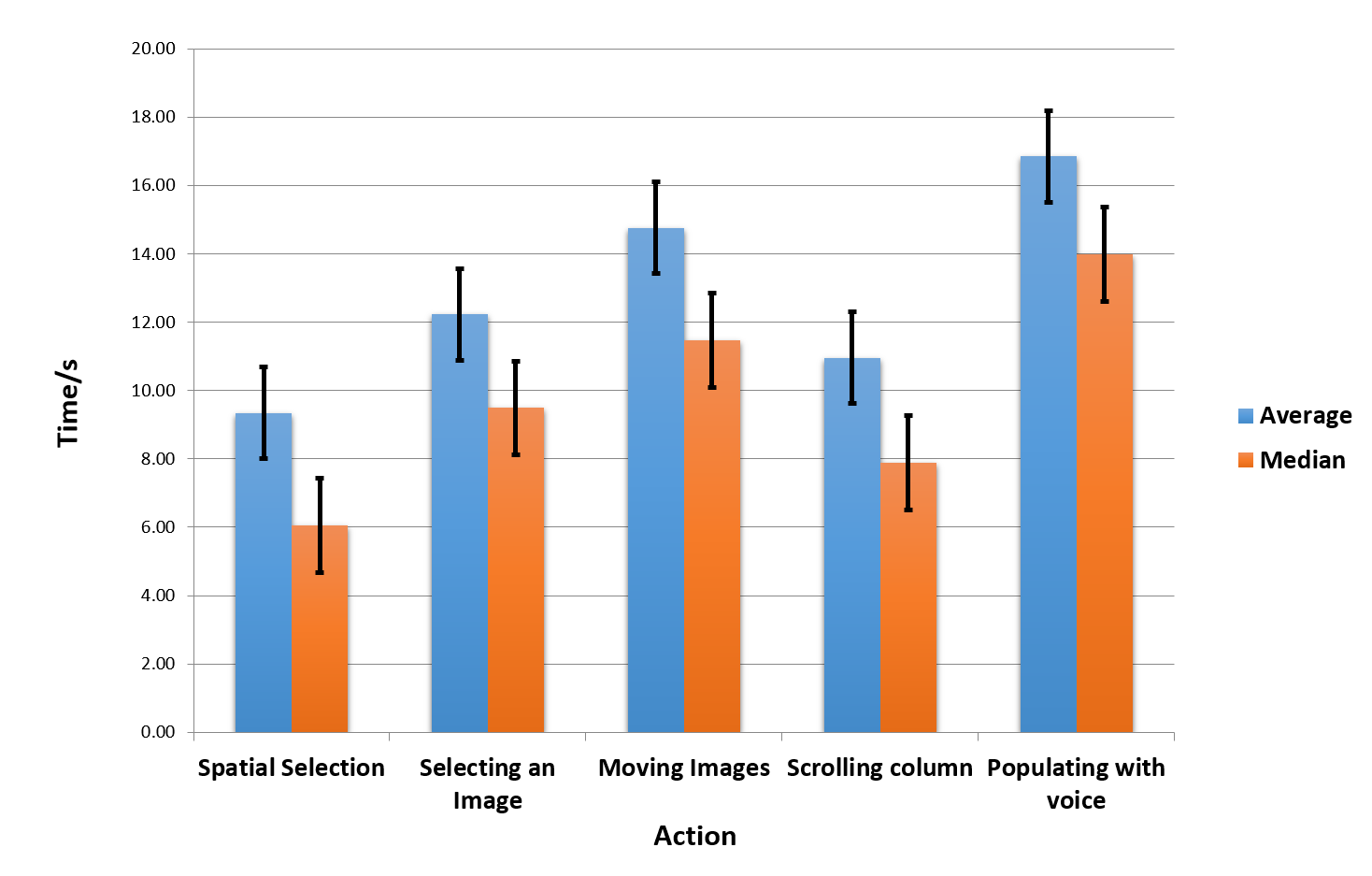}
    \caption{Average and median times for users to complete each action. All actions take longer duration than spatial selection as their completion requires spatial selection as a pre-requisite. %\rnote{This explanation is a little confusing and also maybe something a reviewer coudl pick on.  Also for this and the other bar graphs you should make all the axes label fonts a lot bigger (and probably also the whisker bars thicker).  There shouldn't be a text in the figures that is much smaller than the caption font.}
    }
    \label{fig:sub1}
\end{figure}

Even though we report both the average and median time for each of the actions, we believe that the median times for each of the tasks are more reflective of typical user performance. We observed many cases in which a certain user would take a lot of time to internalize one particular action, while completing other similar actions quickly. This varied significantly from one user to other and therefore led to some higher average values than expected. Unsurprisingly, voice input was the most time consuming action as can be seen in Figure \ref{fig:sub1}.

For Experiment 2, where multiple users worked simultaneously on their personal screens and came together on the shared screen space to complete the full task, we recorded the time of completion. Since there were 3 games for the touch-only interface and 3 for the voice interface, each pair of participants played 6 games. Out of the 21 games for each type of input (7 participant pairs $\times$ 3 games per input type), participants completed 17 of each. 4 games for each input were not completed for various reasons, typically a system crash or one of the participants taking too long to figure out the answer and giving up. The average time taken for a pair of participants to complete the touch-only based game and voice-based game were 2.31 minutes and 1.67 minutes respectively.  Even though experiment 1 revealed that voice input generally takes longer, we note that for touch input the user has to physically move and search for the correct image among a wide array of choices, while for the voice input, users can quickly generate for pictures of their guessed ingredient and move one to the shared screen area.

\begin{figure}[htbp!]
    \centering
    \includegraphics[width=1.0\linewidth]{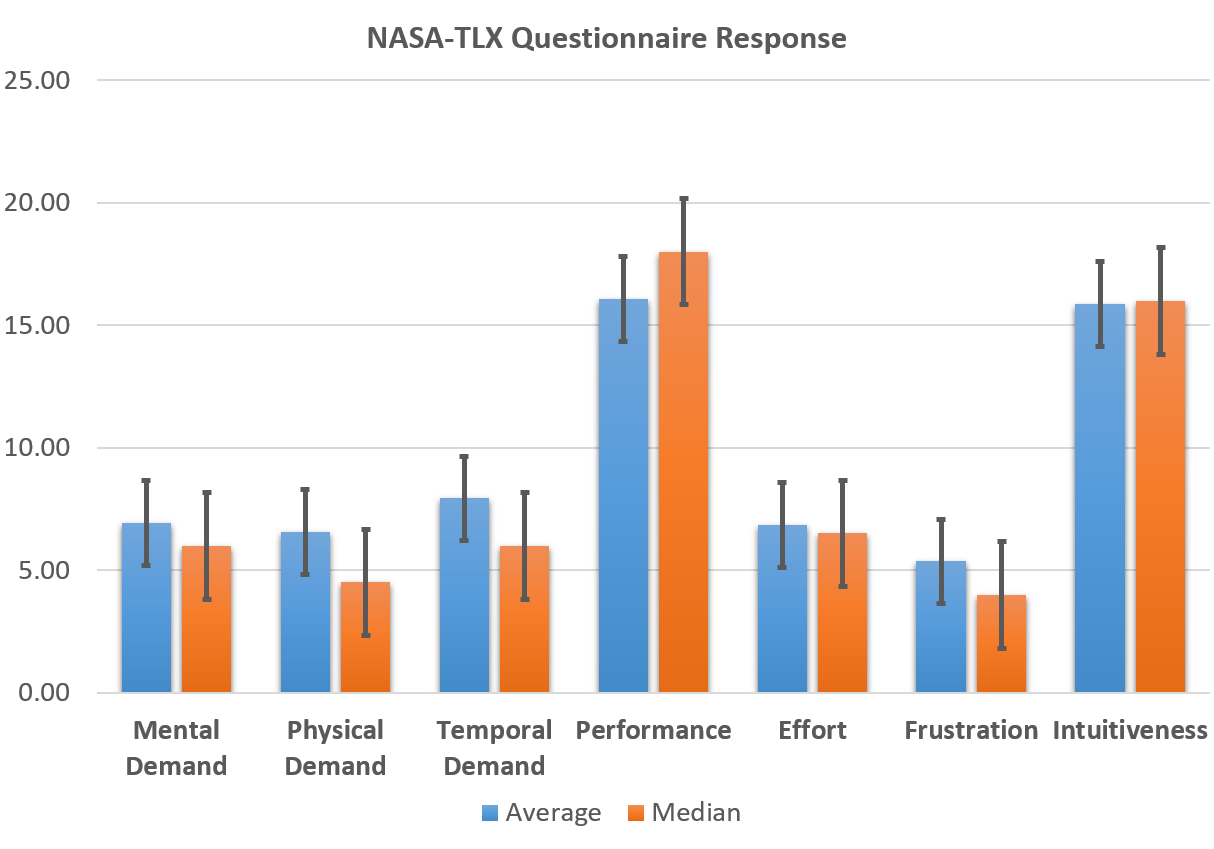}
    \caption{Average and median values of user responses to the NASA-TLX questionnaire.}
    \label{fig:sub2}
\end{figure}

We asked participants to fill out a NASA-TLX questionnaire after completing both experiments to investigate how comfortable and usable our overall system is, and present the results in Figure \ref{fig:sub2}.
We added an extra question regarding the intuitiveness of the overall system, where on the 21 point scale, a higher number indicates a higher degree of intuitiveness. Overall, participants rated their mental, physical, and temporal demand, along with effort and frustration in using the system, to be low. Performance and intuitiveness were highly rated.

\begin{figure}[htbp!]
    \centering
    \includegraphics[width=1.0\linewidth]{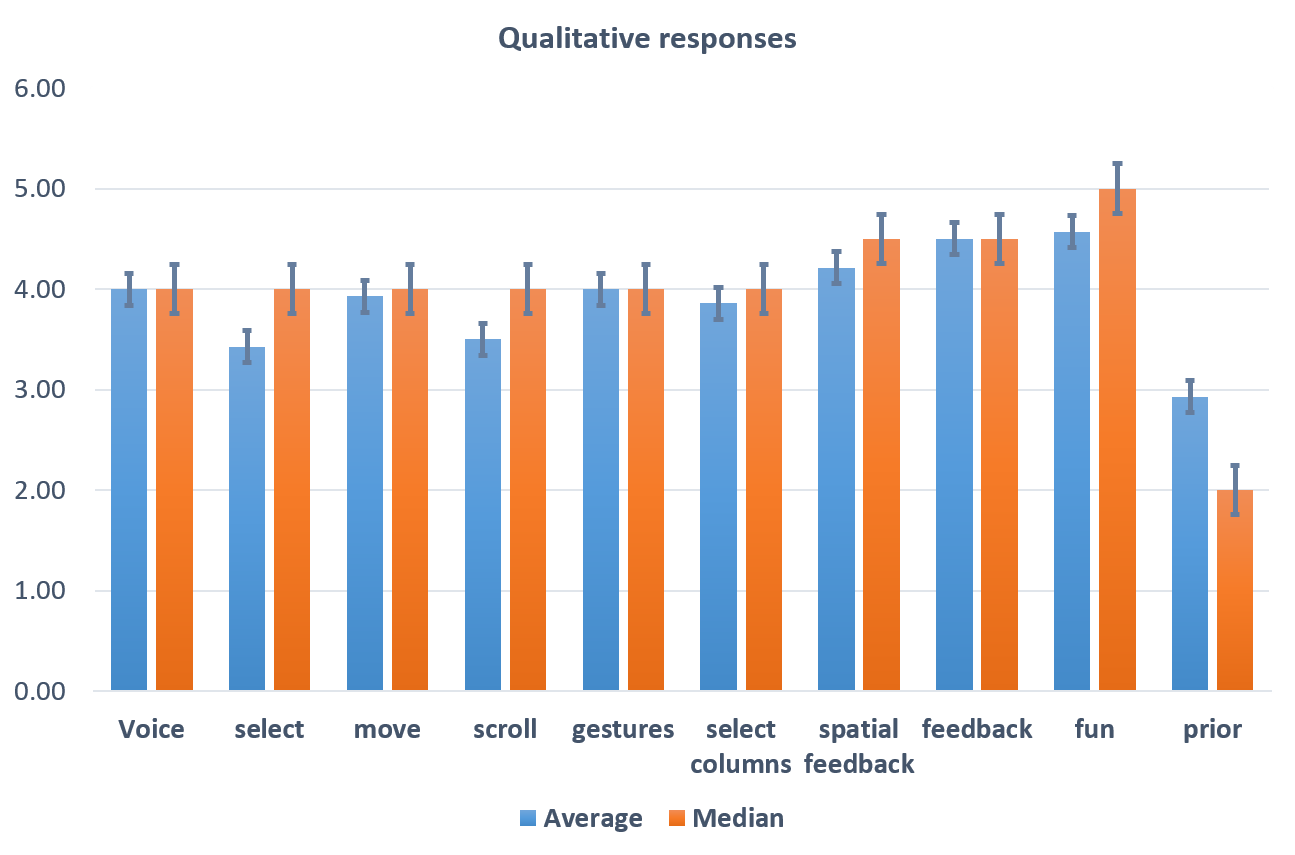}
    \caption{Average and median values of user responses to the second questionnaire on a 5 point Likert scale.}
    \label{fig:sub3}
\end{figure}

In addition, users filled out another questionnaire related to how well they liked/disliked particular interaction mechanisms such as phone gestures, spatial interactions, voice input, and so on, using a 5 point Likert scale. As shown in Figure \ref{fig:sub3}, median values for most of these components are rated very highly. The ratings were also high for whether the overall tasks were fun and enjoyable. Users also highly rated the user interface and other feedback on the screen, including the constant localization feedback.  Among the 14 participants, 3 were previously familiar with the physical space. However, the interaction interface was completely new to them, the same as the rest of the users.  We observed that the users familiar with the large immersive space performed 25\% and 30\% faster than the overall average for the touch and voice games respectively.

We observed that in Experiment 2, the average time for completion with voice input was less than that for the smart phone input, even though Experiment 1 revealed that voice input takes a longer time on average. This can be explained due to the time-consuming nature of search required in the phone input subtask. On the other hand, for the voice input task, upon knowing the key ingredient, users were quickly able to ask for valid pictures and move them to the shared screen area.

Based on observations and post-task interviews, many users appreciated the constant spatial feedback, allowing them to understand their impact on the space. Some users appreciated the automatic demarcation of personal vs.~shared within the scope of the same large screen.

We observed many issues related to the automatic speech understanding and transcription. Non-native English speakers had more difficulty populating their columns with desired input.  Thus it was unsurprising that the average time for actions to be completed using voice was the largest, as shown in Figure \ref{fig:sub1}. Users were divided on the usefulness and comfort of voice input; one user wished he could carry out the entire task using his voice while another was completely opposed to using voice as any kind of input mechanism. 

Many participants gave high marks to the system's approach of mapping the horizontal location of their screen cursor to their physical location and the vertical location to their smartphone screen. However, technical difficulties in which some users had to repeatedly refresh the webpage on their phone due to lost WebSocket connections contributed to a certain level of annoyance.

One of the major usage challenges that many participants commented on was the appropriate appearance of screen prompts and other feedback at eye height. Designing user interfaces/feedback for large displays without blocking screen content is a continuing challenge for this type of research.

\section{Discussion and Future Work}
In terms of overall performance, the results were very encouraging in regards to the usefulness of the overall interaction interface. Using standalone methods such as cross-device platforms or voice only methods have shown limited usability in the past~\cite{sarabadani2018automatic}. However, in our case, we see that users adapt well, when multi-modal inputs; touch screen and voice are used in conjunction with automatic interaction mechanisms based on spatial tracking. 

Large display rich immersive spaces such as the one presented in this work draw significant amount of user attention. So, in designing interaction interfaces that do not overwhelm users, it is important to devise methods that require minimal attention. In this regard, using ubiquitous means such as smartphones, has shown considerable success in our work. Furthermore, allowing users to move freely and using the voice commands selectively, only for content generation, helped users to continuously focus on the screen and the task at hand instead of having to repeatedly glance at the phone screen or manually type input commands.   

The multi-modal user interface presented in this work and its success has led us to work towards building use cases that go well beyond the game play experiments presented in this work. We are working towards building a language learning classroom use case, where students match language characters to images. Image selection and placement tasks based on combination of spatial intelligence, cross-device interaction and voice input in a large immersive space can be re-purposed to support classroom activities, where the room in itself is a teaching tool in contrast to conventional classrooms. The learning outcomes through student feedback and overall success of our interface will be important in furthering interaction design choices going forward.

While we only reported 2-user studies here, our immediate next step is to accommodate 3--6 users simultaneously, to fully realize the potential of our immersive environment as a multi-user space.  In addition to direct extensions of the experiments we discussed here, we are investigating how the screen space for each user can be dynamically defined based on their location rather than constrained to one side of the screen.  

We are also working to replace the worn Bluetooth microphones with an ambient microphone array that uses beamforming, along with the users' known locations, to extract utterances for verbal input.  Finally, we hope to conduct more systemic eye-tracking experiments to explore where the users look on the big screen and how often/under what circumstances they glance down at their phone ``touchpad''.

% BALANCE COLUMNS
\balance{}

% REFERENCES FORMAT
% References must be the same font size as other body text.
\bibliographystyle{SIGCHI-Reference-Format}
\bibliography{main}


\begin{thebibliography}{00}

%%% the Web address.

\ifx \showCODEN    \undefined \def \showCODEN     #1{\unskip}     \fi
\ifx \showDOI      \undefined \def \showDOI       #1{{\tt DOI:}\penalty0{#1}\ }
  \fi
\ifx \showISBNx    \undefined \def \showISBNx     #1{\unskip}     \fi
\ifx \showISBNxiii \undefined \def \showISBNxiii  #1{\unskip}     \fi
\ifx \showISSN     \undefined \def \showISSN      #1{\unskip}     \fi
\ifx \showLCCN     \undefined \def \showLCCN      #1{\unskip}     \fi
\ifx \shownote     \undefined \def \shownote      #1{#1}          \fi
\ifx \showarticletitle \undefined \def \showarticletitle #1{#1}   \fi
\ifx \showURL      \undefined \def \showURL       #1{#1}          \fi

\bibitem{ackad2015wild}
{Christopher Ackad}, {Andrew Clayphan}, {Martin Tomitsch}, {and} {Judy Kay}.
  2015.
\newblock \showarticletitle{An in-the-wild study of learning mid-air gestures
  to browse hierarchical information at a large interactive public display}. In
  {\em Proceedings of the 2015 ACM International Joint Conference on Pervasive
  and Ubiquitous Computing}. ACM, 1227--1238.
\newblock


\bibitem{ackad2016skeletons}
{Christopher Ackad}, {Martin Tomitsch}, {and} {Judy Kay}. 2016.
\newblock \showarticletitle{Skeletons and silhouettes: Comparing user
  representations at a gesture-based large display}. In {\em Proceedings of the
  2016 CHI Conference on Human Factors in Computing Systems}. ACM, 2343--2347.
\newblock


\bibitem{anslow2016collaboration}
{Craig Anslow}, {Pedro Campos}, {and} {Joaquim Jorge}. 2016.
\newblock {\em Collaboration Meets Interactive Spaces}.
\newblock Springer.
\newblock


\bibitem{baldauf2016your}
{Matthias Baldauf}, {Florence Adegeye}, {Florian Alt}, {and} {Johannes Harms}.
  2016.
\newblock \showarticletitle{Your browser is the controller: advanced web-based
  smartphone remote controls for public screens}. In {\em Proceedings of the
  5th ACM International Symposium on Pervasive Displays}. ACM, 175--181.
\newblock


\bibitem{ballagas2006smart}
{Rafael Ballagas}, {Jan Borchers}, {Michael Rohs}, {and} {Jennifer~G Sheridan}.
  2006.
\newblock \showarticletitle{The smart phone: a ubiquitous input device}.
\newblock {\em IEEE Pervasive Computing\/} 1 (2006), 70--77.
\newblock


\bibitem{ballendat2010proxemic}
{Till Ballendat}, {Nicolai Marquardt}, {and} {Saul Greenberg}. 2010.
\newblock \showarticletitle{Proxemic interaction: designing for a proximity and
  orientation-aware environment}. In {\em ACM International Conference on
  Interactive Tabletops and Surfaces}. ACM, 121--130.
\newblock


\bibitem{bragdon2011code}
{Andrew Bragdon}, {Rob DeLine}, {Ken Hinckley}, {and} {Meredith~Ringel Morris}.
  2011.
\newblock \showarticletitle{Code space: touch+ air gesture hybrid interactions
  for supporting developer meetings}. In {\em Proceedings of the ACM
  International Conference on Interactive Tabletops and Surfaces}. ACM,
  212--221.
\newblock


\bibitem{brudy2014anyone}
{Frederik Brudy}, {David Ledo}, {Saul Greenberg}, {and} {Andreas Butz}. 2014.
\newblock \showarticletitle{Is anyone looking? mitigating shoulder surfing on
  public displays through awareness and protection}. In {\em Proceedings of The
  International Symposium on Pervasive Displays}. ACM, 1.
\newblock


\bibitem{chenAirTouch}
{Xiang'Anthony Chen} {and} {others}. 2014.
\newblock \showarticletitle{Air+ touch: interweaving touch and in-air
  gestures}. In {\em Proceedings of the 27th annual ACM symposium on User
  interface software and technology}. ACM.
\newblock


\bibitem{cruz-neira1}
{Carolina Cruz-Neira}, {Daniel J.~Sandin}, {Thomas A.~Defanti}, {Robert
  V.~Kenyon}, {and} {John C.~Hart}. 1992.
\newblock \showarticletitle{The CAVE: Audio Visual Experience Automatic Virtual
  Environment}.
\newblock {\em Communication of the ACM\/}  {35} (1992), 64--72.
\newblock
Issue 6.


\bibitem{cruz-neira}
{Carolina Cruz-Neira}, {Jason Leigh}, {Michael Papka}, {Craig Barnes}, {Steven
  M.~Cohen}, {Sumit Das}, {and} {others}. 1993.
\newblock \showarticletitle{Scientist in Wonderland: A Report on Visualization
  Applications in the CAVE Virtual Reality Environment}. IEEE Symposium on
  Research Frontiers in Virtual Reality, 59--66.
\newblock


\bibitem{doshi2017stickyschedule}
{Vishal Doshi}, {Sneha Tuteja}, {Krishna Bharadwaj}, {Davide Tantillo}, {Thomas
  Marrinan}, {James Patton}, {and} {G~Elisabeta Marai}. 2017.
\newblock \showarticletitle{StickySchedule: an interactive multi-user
  application for conference scheduling on large-scale shared displays}. In
  {\em Proceedings of the 6th ACM International Symposium on Pervasive
  Displays}. ACM, 2.
\newblock


\bibitem{du2011tilt}
{Yuan Du}, {Haoyi Ren}, {Gang Pan}, {and} {Shjian Li}. 2011.
\newblock \showarticletitle{Tilt \& touch: Mobile phone for 3D interaction}. In
  {\em Proceedings of the 13th international conference on Ubiquitous
  computing}. ACM, 485--486.
\newblock


\bibitem{febretti2013cave2}
{Alessandro Febretti}, {Arthur Nishimoto}, {Terrance Thigpen}, {Jonas
  Talandis}, {Lance Long}, {JD Pirtle}, {Tom Peterka}, {Alan Verlo}, {Maxine
  Brown}, {Dana Plepys}, {and} {others}. 2013.
\newblock \showarticletitle{CAVE2: a hybrid reality environment for immersive
  simulation and information analysis}. In {\em IS\&T/SPIE Electronic Imaging}.
  International Society for Optics and Photonics, 864903--864903.
\newblock


\bibitem{hawkey2005proximity}
{Kirstie Hawkey}, {Melanie Kellar}, {Derek Reilly}, {Tara Whalen}, {and}
  {Kori~M Inkpen}. 2005.
\newblock \showarticletitle{The proximity factor: impact of distance on
  co-located collaboration}. In {\em Proceedings of the 2005 international ACM
  SIGGROUP conference on Supporting group work}. ACM, 31--40.
\newblock


\bibitem{horak2016presenting}
{Tom Horak}, {Ulrike Kister}, {and} {Raimund Dachselt}. 2016.
\newblock \showarticletitle{Presenting business data: Challenges during board
  meetings in multi-display environments}. In {\em Proceedings of the 2016 ACM
  International Conference on Interactive Surfaces and Spaces}. ACM, 319--324.
\newblock


\bibitem{kister2017grasp}
{Ulrike Kister}, {Konstantin Klamka}, {Christian Tominski}, {and} {Raimund
  Dachselt}. 2017.
\newblock \showarticletitle{GraSp: Combining Spatially-aware Mobile Devices and
  a Display Wall for Graph Visualization and Interaction}. In {\em Computer
  Graphics Forum}, Vol.~36. Wiley Online Library, 503--514.
\newblock


\bibitem{kister2016multilens}
{Ulrike Kister}, {Patrick Reipschl{\"a}ger}, {and} {Raimund Dachselt}. 2016.
\newblock \showarticletitle{Multilens: Fluent interaction with multi-functional
  multi-touch lenses for information visualization}. In {\em Proceedings of the
  2016 ACM International Conference on Interactive Surfaces and Spaces}. ACM,
  139--148.
\newblock


\bibitem{kopper2008increasing}
{Regis Kopper}, {Mara~G Silva}, {Ryan~Patrick McMahan}, {and} {Doug~A Bowman}.
  2008.
\newblock \showarticletitle{Increasing the precision of distant pointing for
  large high-resolution displays}.
\newblock  (2008).
\newblock


\bibitem{kropp2017enhancing}
{Martin Kropp}, {Craig Anslow}, {Magdalena Mateescu}, {Roger Burkhard}, {Dario
  Vischi}, {and} {Carmen Zahn}. 2017.
\newblock \showarticletitle{Enhancing Agile Team Collaboration Through the Use
  of Large Digital Multi-touch Cardwalls}. In {\em International Conference on
  Agile Software Development}. Springer, Cham, 119--134.
\newblock


\bibitem{lander2015collaborative}
{Christian Lander}, {Marco Speicher}, {Denise Paradowski}, {Norine Coenen},
  {Sebastian Biewer}, {and} {Antonio Kr{\"u}ger}. 2015.
\newblock \showarticletitle{Collaborative newspaper: Exploring an adaptive
  scrolling algorithm in a multi-user reading scenario}. In {\em Proceedings of
  the 4th International Symposium on Pervasive Displays}. ACM, 163--169.
\newblock


\bibitem{langner2019multiple}
{Ricardo Langner}, {Ulrike Kister}, {and} {Raimund Dachselt}. 2019.
\newblock \showarticletitle{Multiple Coordinated Views at Large Displays for
  Multiple Users: Empirical Findings on User Behavior, Movements, and
  Distances}.
\newblock {\em IEEE transactions on visualization and computer graphics\/}
  {25}, 1 (2019), 608--618.
\newblock


\bibitem{liu2014leveraging}
{Can Liu}. 2014.
\newblock \showarticletitle{Leveraging physical human actions in large
  interaction spaces}. In {\em Proceedings of the adjunct publication of the
  27th annual ACM symposium on User interface software and technology}. ACM,
  9--12.
\newblock


\bibitem{luojus2013wordster}
{Petri Luojus}, {Jarkko Koskela}, {Kimmo Ollila}, {Saku-Matti M{\"a}ki}, {Raffi
  Kulpa-Bogossia}, {Tommi Heikkinen}, {and} {Timo Ojala}. 2013.
\newblock \showarticletitle{Wordster: collaborative versus competitive gaming
  using interactive public displays and mobile phones}. In {\em Proceedings of
  the 2nd ACM International Symposium on Pervasive Displays}. ACM, 109--114.
\newblock


\bibitem{malik2005interacting}
{Shahzad Malik}, {Abhishek Ranjan}, {and} {Ravin Balakrishnan}. 2005.
\newblock \showarticletitle{Interacting with large displays from a distance
  with vision-tracked multi-finger gestural input}. In {\em Proceedings of the
  18th annual ACM symposium on User interface software and technology}. ACM,
  43--52.
\newblock


\bibitem{nutsi2015usability}
{Andrea Nutsi}. 2015.
\newblock \showarticletitle{Usability Guidelines for Co-Located Multi-User
  Interaction on Wall Displays}. In {\em Proceedings of the 2015 International
  Conference on Interactive Tabletops \& Surfaces}. ACM, 433--438.
\newblock


\bibitem{nutsi2015multi}
{Andrea Nutsi} {and} {Michael Koch}. 2015.
\newblock \showarticletitle{Multi-User Usability Guidelines for Interactive
  Wall Display Applications}. In {\em Proceedings of the 4th International
  Symposium on Pervasive Displays}. ACM, 233--234.
\newblock


\bibitem{rittenbruch2013cube}
{Markus Rittenbruch}, {Andrew Sorensen}, {Jared Donovan}, {Debra Polson},
  {Michael Docherty}, {and} {Jeff Jones}. 2013.
\newblock \showarticletitle{The cube: a very large-scale interactive engagement
  space}. In {\em Proceedings of the 2013 ACM international conference on
  Interactive tabletops and surfaces}. ACM, 1--10.
\newblock


\bibitem{sarabadani2018automatic}
{Amir~E Sarabadani~Tafreshi}, {Andrea Soro}, {and} {Gerhard Tr{\"o}ster}. 2018.
\newblock \showarticletitle{Automatic, Gestural, Voice, Positional, or
  Cross-Device Interaction? Comparing Interaction Methods to Indicate Topics of
  Interest to Public Displays}.
\newblock {\em Frontiers in ICT\/}  {5} (2018), 20.
\newblock


\bibitem{gsharma}
{Gyanendra Sharma}, {Jonas Braasch}, {and} {Richard~J. Radke}. 2016.
\newblock \showarticletitle{Interactions in a Human-Scale Immersive
  Environment: the {CRAIVE-Lab}}. In {\em Cross-Surface Workshop at ISS2016}.
  Niagara Falls, Canada.
\newblock


\bibitem{shirazi2009flashlight}
{Alireza~Sahami Shirazi}, {Christian Winkler}, {and} {Albrecht Schmidt}. 2009.
\newblock \showarticletitle{Flashlight interaction: a study on mobile phone
  interaction techniques with large displays}. In {\em Proceedings of the 11th
  International Conference on Human-Computer Interaction with Mobile Devices
  and Services}. ACM, 93.
\newblock


\bibitem{vogel2004interactive}
{Daniel Vogel} {and} {Ravin Balakrishnan}. 2004.
\newblock \showarticletitle{Interactive public ambient displays: transitioning
  from implicit to explicit, public to personal, interaction with multiple
  users}. In {\em Proceedings of the 17th annual ACM symposium on User
  interface software and technology}. ACM, 137--146.
\newblock


\bibitem{von2017giant}
{Ulrich von Zadow} {and} {Raimund Dachselt}. 2017.
\newblock \showarticletitle{Giant: Visualizing group interaction at large wall
  displays}. In {\em Proceedings of the 2017 CHI Conference on Human Factors in
  Computing Systems}. ACM, 2639--2647.
\newblock


\bibitem{wallace2017subtle}
{James~R Wallace}, {Ariel Weingarten}, {and} {Edward Lank}. 2017.
\newblock \showarticletitle{Subtle and Personal Workspace Requirements for
  Visual Search Tasks on Public Displays}. In {\em Proceedings of the 2017 CHI
  Conference on Human Factors in Computing Systems}. ACM, 6760--6764.
\newblock


\bibitem{wolf2016proxemic}
{Katrin Wolf}, {Yomna Abdelrahman}, {Thomas Kubitza}, {and} {Albrecht Schmidt}.
  2016.
\newblock \showarticletitle{Proxemic zones of exhibits and their manipulation
  using floor projection}. In {\em Proceedings of the 5th ACM International
  Symposium on Pervasive Displays}. ACM, 33--37.
\newblock


\bibitem{yoo2015dwell}
{Soojeong Yoo}, {Callum Parker}, {Judy Kay}, {and} {Martin Tomitsch}. 2015.
\newblock \showarticletitle{To dwell or not to dwell: an evaluation of mid-air
  gestures for large information displays}. In {\em Proceedings of the Annual
  Meeting of the Australian Special Interest Group for Computer Human
  Interaction}. ACM, 187--191.
\newblock


\end{thebibliography}

\end{document}